\documentclass[runningheads]{llncs}
\usepackage{graphicx}
\usepackage{fixltx2e}
\usepackage[title]{appendix}
\usepackage{multirow}
\usepackage{adjustbox}
\usepackage{array}
\usepackage{perpage} %the perpage package
\MakePerPage{footnote} %the perpage package command
\newcolumntype{P}[1]{>{\centering\arraybackslash}p{#1}}

\begin{document}
\title{SPBERT: An Efficient Pre-training BERT on SPARQL Queries for Question Answering over Knowledge Graphs}
\author{Hieu Tran\inst{1,2} \and
Long Phan\inst{3} \and
James Anibal\inst{4} \and
Binh T. Nguyen\inst{1,2} \and
Truong-Son Nguyen\inst{1,2}}
\authorrunning{Tran et al.}
\titlerunning{SPBERT: A Pre-trained Model for SPARQL Query Language}
\institute{
University of Science, Ho Chi Minh City, Vietnam \and
Vietnam National Univeristy, Ho Chi Minh City, Vietnam \and
Case Western Reserve University, Ohio, USA \and
 National Cancer Institute, Maryland, USA \\
\email{long.phan@case.edu, ntson@fit.hcmus.edu.vn}}
\maketitle            % typeset the header of the contribution
\begin{abstract}

% Knowledge Graphs (KGs) is becoming increasingly popular and necessary during the past years. 
% Unlike Information Retrieval-based, Question Answering (QA) over KGs employs structured data between relational entities stored in knowledge bases to answer natural language related tasks. Recently, the emergence of pre-trained language models such as BERT has stimulated the development of research in natural language understanding. Yet, to the best of our knowledge, there has been no attempt into employing pre-train language model (LM) into generating an end-to-end Question Answering system supporting SPARQL Query Construction or Answer Verbalization Generation from existed Query. In this article, we investigate how the recently introduced pre-trained language model BERT and encoder-decoder architecture can be adapted for Question Answering over Knowledge Graphs (KGs) corpora.

In this paper, we propose SPBERT, a transformer-based language model pre-trained on massive SPARQL query logs. By incorporating masked language modeling objectives and the word structural objective, SPBERT can learn general-purpose representations in both natural language and SPARQL query language. We investigate how SPBERT and encoder-decoder architecture can be adapted for Knowledge-based QA corpora. We conduct exhaustive experiments on two additional tasks, including SPARQL Query Construction and Answer Verbalization Generation. The experimental results show that SPBERT can obtain promising results, achieving state-of-the-art BLEU scores on several of these tasks.

\keywords{Machine translation  \and SPARQL \and Language models \and Question answering.}
\end{abstract}
\section{Introduction}
During the last decade, pre-trained language models (LM) have been playing an essential role in many areas of natural language processing (NLP), including Question Answering (QA) \cite{DBLP:journals/corr/abs-1804-07461}. Large pre-trained models like GPT \cite{brown2020language}, BERT \cite{devlin-etal-2019-bert}, and XLNet \cite{DBLP:journals/corr/abs-1906-08237} derive contextualized word vector representations from immense text corpora. It can represent a significant deviation from traditional word embedding methods wherein each word is given a global representation. After training, one can usually fine-tune the corresponding models for downstream tasks. These pre-trained models have dramatically improved the state-of-the-art performance on multiple downstream NLP tasks. As such, this concept has been extended into various domains. One can retrain BERT models on corpora containing text specific to a particular area; for instance, Feng and colleagues used CodeBERT for tasks involving programming languages \cite{feng-etal-2020-codebert}.

There has also been widespread use of publicly available Knowledge Graph (KG) datasets, such as DBpedia\footnote{https://dbpedia.org}, Wikidata\footnote{https://wikidata.org}, or Yago\footnote{https://yago-knowledge.org}. These datasets provide a valuable source of structured knowledge for NL relational tasks, including QA. These knowledge base (KB) endpoints use a query language called SPARQL, a standard graphing-matching query language to query graph data in the form of RDF triples \cite{Manola_04} (an example format of a SPARQL query can be referred to Table \ref{example_table}).  
One of the key challenges in these SPAQRL-NL generation tasks is understanding the structured schemas between entities and relations in the KGs. In addition, it is necessary to correctly generate NL answers from SPARQL queries and generate SPARQL queries from NL descriptions. These challenges call for the development of pre-trained language models (LMs) that understand SPARQL schemas' structure in KGs and free-form text. 
While existing LMs (e.g., BERT) are only trained on encoding NL text, we present an unprecedented approach for developing an LM that supports NL and SPARQL query language (SL).

Previous works in this area \cite{DBLP:journals/corr/abs-1906-09302} utilized the traditional word embedding method and built the vocabulary from tokens separated by spaces. The higher the number of entities and predicates in the dataset, the larger the vocabulary size. It is a notable drawback that needs to be improved to reduce time and cost complexity. We present SPBERT, a pre-trained model for SPARQL query language (SL) using a common vocabulary. SPBERT is a multi-layer transformer-based model \cite{DBLP:journals/corr/VaswaniSPUJGKP17}, in which the architecture has been proven effective for various other large LMs models. We train SPBERT using standard masked language modeling \cite{devlin-etal-2019-bert}. To further advance the exhibit of the underlying solid structure of SPARQL schemas, we also incorporate the word structural objective \cite{DBLP:journals/corr/abs-1908-04577}. These structural learning objectives can enable SPBERT to gain insights into the word-level structure of SL language during pre-training. 
We train SPBERT from large, processed SPARQL query logs. To show the effectiveness of our approach in creating a general-purpose pre-trained LMs for NL and SL, SPBERT is fine-tuned and evaluated on six popular datasets in SPARQL query construction and answer verbalization generation topics. In addition, we test various pre-training strategies, including different learning objectives and model checkpoints.

Our paper provides the following contributions: (1) We investigate how pre-trained language models can be applied to the sequence-to-sequence architecture in KBQA; (2) We introduce SPBERT, the first pre-trained language model for SPARQL Query Language that focuses on understanding the query structure; (3) We show that pre-training on large SPARQL query corpus and incorporating word-level ordering learning objectives lead to better performance. Our model achieves competitive results on Answer Verbalization Generation and SPARQL Query Construction; (4) We make our pre-trained checkpoints of SPBERT and related source code\footnote{https://github.com/heraclex12/NLP2SPARQL} for fine-tuning publicly available.

\section{Related Work}
BERT \cite{devlin-etal-2019-bert} is a pre-trained contextualized word representation model which consists of the encoder block taken from the transformer method. \cite{DBLP:journals/corr/VaswaniSPUJGKP17}. BERT is trained with masked language modeling to learn word representation in both left and right contexts. The model incorporates information from bidirectional representations in a sequence - this has proved to be effective in learning a natural language. We hypothesize that this architecture will be effective in more domain-specific SPARQL query language contexts.

Multiple prior attempts use deep learning for enhancing performance on natural language-SPARQL query language-related tasks. For example, Yin et al. \cite{DBLP:journals/corr/abs-1906-09302} presented different experiments with three types of neural machine translation models, including RNN-based, CNN-based, and Transformer-based models for translating natural language to SPARQL query tasks. The results showed a dominance of the convolutional sequence-to-sequence model over all of the proposed models across five datasets. This method is highly correlated with our approach, so we treat this work as our baseline later in this paper.
Luz and Finger \cite{DBLP:journals/corr/abs-1803-04329} proposed an LSTM encoder-decoder model that is capable of encoding natural language and decoding the corresponding SPARQL query. Furthermore, this work presented multiple methods for generating vector representation of Natural Language and SPARQL language. Finally, the paper introduced a novel approach for developing a lexicon representation vector of SPARQL. The results illustrated that this approach could achieve state-of-the-art results on natural language-SPARQL datasets.

There have also been attempts to developing a Knowledge-based question answering (KBQA) system that leverages multiple models for the various tasks. For example, Kapanipathi and colleagues introduced semantic parsing and reasoning-based Neuro-Symbolic Question Answering (NSQA) system \cite{kapanipathi2020question}. The model consisted of an Abstract Meaning Representation (AMR) layer that could parse the input question, a path-based approach to transform the AMR into KB logical queries, and a Logical Neural Network (LNN) to filter out invalid queries. As a result, NSQA could achieve state-of-the-art performance on KBQA datasets.
\section{SPBERT}

\subsection{Pre-training Data} \label{pretrainingdata}
To prepare a large-scale pre-training corpus, we leverage SPARQL queries from end-users, massive and highly diverse structures. These query logs can be obtained from the DBpedia endpoint\footnote{https://dbpedia.org/sparql} powered by a Virtuoso\footnote{https://virtuoso.openlinksw.com} instance. We only focus on valid DBpedia query logs spans from October 2015 to April 2016.
These raw queries contain many duplicates, arbitrary variable names, and unnecessary components such as prefixes and comments. 

To address these issues, we prepared a heuristics pipeline to clean the DBpedia SPARQL query logs and released this pipeline with the model\footnotemark[4]. This pipeline includes a cleaning process specific to SPARQL query language (i.e., removing comments, excluding prefix declarations, converting namespace, filtering unknown links, standardizing white space and indentations, etc.). In addition, the processed queries will go through an encoding process suggested by \cite{DBLP:journals/corr/abs-1906-09302} to make these queries look more natural.
We obtained approximately 6.8M queries. An example SPARQL query can be depicted in Table \ref{example_table}.        
\def\arraystretch{1.2}% 
\setlength{\tabcolsep}{0.9em} %

\begin{table}[htbp]
\centering
\caption{An example of SPARQL query and the corresponding encoding}
\begin{tabular}{p{2.2cm}|p{8.8cm}}
\hline
Original query & SELECT DISTINCT ?uri \\
& WHERE \{ $<$http://dbpedia.org/resource/Tom\_Hanks$>$ $<$http://dbpedia.org/ontology/spouse$>$ ?uri \} \\ 
\hline
Encoded query &  select distinct var\_uri where brack\_open $<$dbr\_Tom\_Hanks$>$ $<$dbo\_spouse$>$ var\_uri brack\_close \\
\hline
\end{tabular}
\label{example_table}

\end{table}
% Our goal is to develop the models that can learn general-purpose SPARQL query language representations and transfer this knowledge to the tasks of understanding SPARQL syntax. In this section, we briefly discuss how to preprocess the SPARQL query corpus for pre-training, and then we introduce the input representation, what the model looks like, and how to fine-tune our proposed models on natural language-SPARQL query language tasks.

\subsection{Input Representation}
The input of SPBERT is a sequence of tokens of a single, encoded SPARQL query. We follow the style of input representation used in BERT. Every input sequence requires a special classification token, \emph{[CLS]} as the first token, and a special end-of-sequence, \emph{[SEP]}. \emph{[CLS]} token contains representative information of the whole input sentence, while the \emph{[SEP]} token is used to separate different sentences of an input. To alleviate the out-of-vocabulary problem in tokenization, we use WordPiece \cite{wu2016googles} to split the query sentence into subword units.
In addition, we employ the same vocabulary with cased BERT for the following reasons: (i) SPARQL queries are almost made up of English words, and we aim to leverage existing pre-trained language models; (ii) these queries require strictly correct representations of entities and relations, which can contain either lowercase and uppercase characters. All tokens of the input sequence can be mapped to this vocabulary to pick up their corresponding indexes and then feed these indexes into the model.

\subsection{Model Architecture}
\begin{figure}
  \centering
  \includegraphics[width=0.81\textwidth]{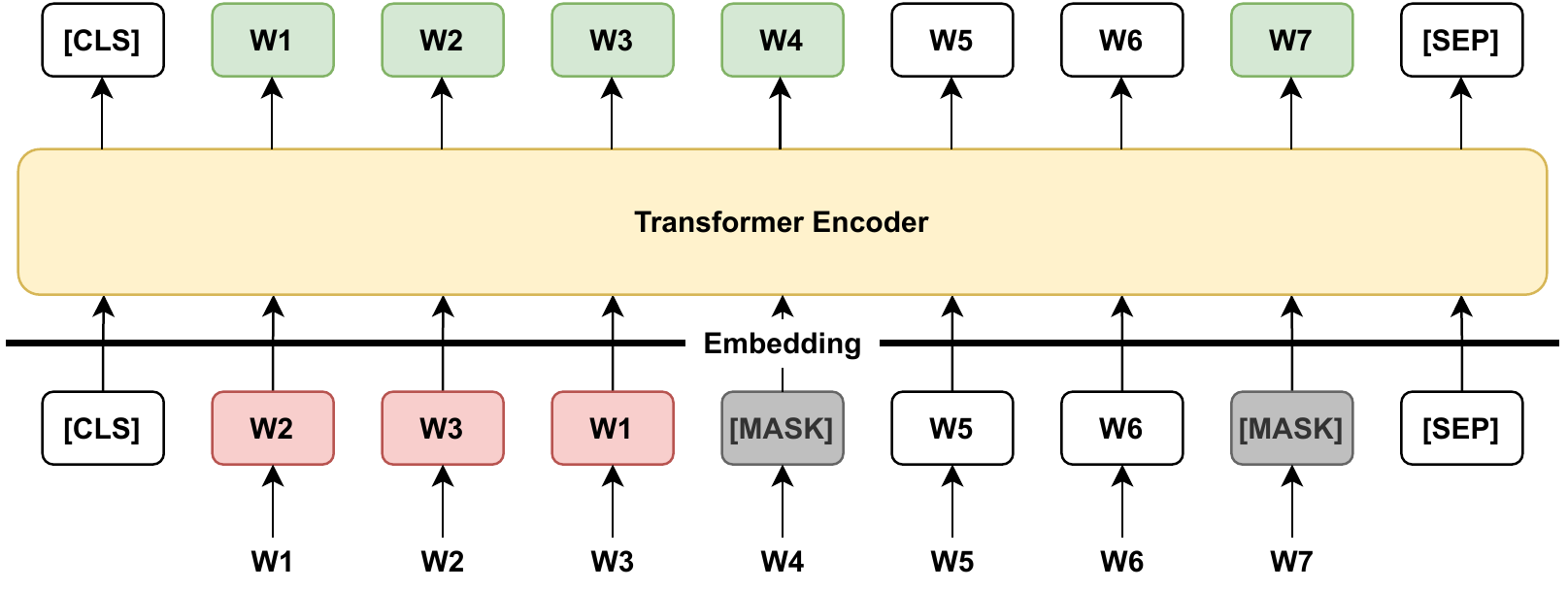}
  \caption{An illustration about the combination of MLM and WSO}
  \label{fig:model}
\end{figure}

SPBERT uses the same architecture as BERT \cite{devlin-etal-2019-bert}, which is based on a multi-layer bidirectional transformer \cite{DBLP:journals/corr/VaswaniSPUJGKP17}. BERT model is trained on two auxiliary tasks: masked language modeling and next sentence prediction. As our training corpus only contains independent SPARQL queries, we substitute the next sentence prediction with the word structural objective \cite{DBLP:journals/corr/abs-1908-04577}. We illustrate how to combine these two tasks in Figure \ref{fig:model}.
\subsubsection{Task \#1: Masked Language Modeling (MLM)} randomly replaces some percentage of tokens from a sequence (similar to \cite{devlin-etal-2019-bert}, the rate of 15\% of the tokens will be covered) with a special token [MASK]. The model will then try to reconstruct this sequence by predicting the original token.

\subsubsection{Task \#2: Word Structural Objective (WSO)} gives our model the ability to capture the sequential order dependency of a sentence. This objective corrupts the right order of words by randomly selecting n-grams from unmasked tokens and then permuting the elements. The model has to predict the original order of tokens to correct the sentence. Different from \cite{DBLP:journals/corr/abs-1908-04577}, our pre-training data are mostly short queries (less than 256 tokens). Therefore, we permute 10\% of the n-grams rather than 5\%.

We released three different versions of SPBERT. The first version begins training with randomly initialized weights, and the second starts with the weights from the pre-trained BERT model. Both these models use only the MLM objective, while we combine MLM and WSO to train the third SPBERT.

\subsection{Pre-training Setup}
In the pre-training step, we denote the number of Transformer encoder layers as L, the size of hidden vectors as H, and the number of self-attention heads as A. We followed the setting of BERT\textsubscript{BASE} (L=12, H=768, A=12, total parameters=110M) and continued to train 200K steps from cased BERT\textsubscript{BASE} checkpoint. The maximum sequence length was fixed to 512, and the batch size was set to 128. We used Adam with a learning rate of 2e-5 and epsilon of 1e-8 and employed cased BERT\textsubscript{BASE} vocabulary with 30K tokens. 

\subsection{Fine-tuning SPBERT}
Transformer-based sequence-to-sequence architecture \cite{rothe-etal-2020-leveraging} is an encoder-decoder architecture that employs multi-layer self-attention to efficiently parallelize long-term dependencies. This architecture has two components: an encoder and a decoder. The encoder can capture a contextualized representation of input sequences, and the decoder uses this information to generate target sequences. Intuitively, this architecture is often used for generating a target sequence from a source sequence. 

To fine-tune SPBERT on the natural language-SPARQL query language tasks required for an end-to-end KBQA, we apply pre-trained language models to the Transformer-based sequence-to-sequence architecture. BERT and SPBERT are encoder-only models developed for encoding language representations, so we can quickly assemble this model into the encoder. However, to adapt BERT-based models to the decoder, we must change the self-attention layers from bidirectional to left-context-only. We must also insert a cross-attention mechanism with random weights. For tasks with the input as a question and the output as a query, we initialize the encoder with BERT checkpoints and the decoder with SPBERT checkpoints. We denote this model as BERT2SPBERT.
Conversely, when we use SPARQL queries as input and NL answers are the outputs, SPBERT will be initialized as the encoder. The decoder can be initialized at random (SPBERT2RND) or from BERT checkpoints (SPBERT2BERT). Similar to the pre-training phase, all SPARQL queries will be encoded before being put into the model. 

We apply the weight typing method to reduce the number of parameters but still boost the performance. Same as \cite{press-wolf-2017-using}, we tie the input embedding and the output embedding of the decoder that help the output embedding makes use of the weights learned from the input embedding instead of initializing randomly.

\section{Experiments}
This section first describes the datasets used to evaluate our proposed models and then explains the experimental setup and the results.

\subsection{Datasets}\label{dataset}
An end-to-end KBQA system must perform two tasks sequentially. First, this system acquires a question and constructs a corresponding SPARQL query. We refer to this task as SPARQL Query Construction. The remaining task is Answer Verbalization Generation which inputs the generated queries from the previous one and produces natural language answers. In these experiments, we only consider English datasets, and all queries can be encoded in the way described in Section \ref{pretrainingdata}. We represent both tasks below and summarize the evaluation datasets in Table \ref{statistic_table}. 

\subsubsection{SPARQL Query Construction}
QALD-9 (Question Answering over Linked Data) \cite{Usbeck20189thCO} consists of a question and query pairs from real-world questions and query logs.
LC-QuAD (Large-scale Complex Question Answering Dataset) \cite{10.1007/978-3-319-68204-4_22}
contains 5000 questions and corresponding queries from the DBPedia dataset. Each NL question in the LC-QuAD dataset is peer-reviewed by fluent English speakers to ensure the quality of the data.
\def\arraystretch{1.2}% 
\setlength{\tabcolsep}{0.9em} %

\begin{table}[htbp]
\centering
\caption{Data statistics about Question Answering datasets over Knowledge Graphs}
\begin{tabular}{p{2.2cm}|P{1.2cm}|P{1.4cm}|P{1.1cm}|P{1.1cm}|p{1.3cm}}
\hline
\multirow{2}{0pt}{\centering\textbf{Dataset}} & \multicolumn{3}{c|}{\textbf{Amount}} & \multirow{2}{1.1cm}{\centering\textbf{Tokens}} & \multirow{2}{0pt}{\centering\textbf{Creation}} \\
\cline{2-4}
 & Training & Validation & Test &  \\
\hline
QALD-9 \cite{Usbeck20189thCO} & 408 & - & 150 & 1042 & Manual \\
LC-QuAD \cite{10.1007/978-3-319-68204-4_22} & 4000 & 500 & 500 & 5035 & Manual \\
Mon \cite{soru2020sparql} & 14588 & 100 & 100 & 2066 & Automatic \\
Mon50 \cite{DBLP:journals/corr/abs-1906-09302} & 7394 & 1478 & 5916 & 2066 & Automatic \\
Mon80 \cite{DBLP:journals/corr/abs-1906-09302} & 11830 & 1479 & 1479 & 2066 & Automatic \\
\hline
\hline
VQuAnDa \cite{10.1007/978-3-030-49461-2_31} & 4000 & 500 & 500 & 5035 & Manual \\
\hline
\end{tabular}
\label{statistic_table}

\end{table}
The Mon dataset was introduced by \cite{soru2020sparql}. This dataset contains 38 pairs of handcrafted question and query templates. Each template is automatically inserted into one or two entities of the Monument ontology. In order to compare the performance between different number of training samples, \cite{DBLP:journals/corr/abs-1906-09302} splits this dataset using two different ratios for training, test, and validation sets 80\%-10\%-10\% (Mon80) and 50\%-10\%-40\% (Mon50).

\subsubsection{Answer Verbalization Generation}
VQuAnDa (Verbalization QUestion ANswering DAtaset) \cite{10.1007/978-3-030-49461-2_31} extends the LC-QuAD dataset by providing the query answers in natural language. These answers are created based on the questions and the corresponding SPARQL queries. An automatic framework generates the templates for the answers and uses a rule-based method to produce the first version. The final results are manually reviewed and rephrased to ensure grammatical correctness. We give a sample of this dataset in Table \ref{data_example}.
\def\arraystretch{1.2}% 
\setlength{\tabcolsep}{0.9em} %

\begin{table}[htbp]
\centering
\caption{An example of two KBQA tasks}
\begin{tabular}{|p{1.5cm}|p{9.5cm}|}
\hline
Question & How many people play for the Dallas Cowboys? \\
\hline
\multicolumn{2}{c}{\textbf{With Entites}} \\
\hline
SPARQL &  select distinct count(var\_uri) where brack\_open var\_uri $<$dbo\_team$>$ $<$dbr\_Dallas\_Cowboys$>$ brack\_close \\
\hline
Answer &  $<$ans$>$ people play for the dallas cowboys . \\
\hline
\multicolumn{2}{c}{\textbf{Covered Entities}} \\
\hline
SPARQL &  select distinct count(var\_uri) where brack\_open var\_uri $<$dbo\_team$>$ $<$ent$>$ brack\_close \\
\hline
Answer &  $<$ans$>$ people play for the $<$ent$>$ . \\
\hline
\end{tabular}
\label{data_example}

\end{table}

\subsection{Experimental Setup}
We verify the effectiveness of our proposed methods by comparing these models with three following types of network architectures from previous works \cite{10.1007/978-3-030-49461-2_31,DBLP:journals/corr/abs-1906-09302}.
\begin{itemize}
 \item \textbf{RNN-based models}: These models are based on a standard sequence-to-sequence architecture and combine variants of RNNs (such as Long Short-Term Memory and Gated Recurrent Unit) with an attention mechanism.
 \item \textbf{CNN-based models}: CNN-based sequence to sequence (ConvS2S) models, which leverage an encoder-decoder architecture with an attention mechanism. In these cases, both the encoder and decoder consist of stacked convolutional layers.  
 \item \textbf{Transformer models}: These models are based on \cite{DBLP:journals/corr/VaswaniSPUJGKP17}, in which each layer of encoder-decoder architecture includes two major components: a multi-head self-attention layer and a feed-forward network. Since these models are initialized with random weights, they also are called RND2RND.
\end{itemize}

\subsubsection{Performance metrics}
The most reliable evaluation method is to have appropriately qualified experts look at the translation and evaluate manually. However, this evaluation is costly and time-consuming. 
Therefore, we are using automatic metrics that are faster and cheaper but still correlate with human judgments. We use corpus level Bilingual Evaluation Understudy (BLEU) \cite{papineni-etal-2002-bleu} and Exact Match score (EM) as follows:
(1) \textbf{BLEU }measure the differences in word choice and word order between candidate translations and reference translations by calculating the overlap of n-grams between a generated sequence and one or more reference sequences.
(2) \textbf{EM} is the ratio of the number of generated sequences that perfectly match reference to the total number of input samples. 

\subsubsection{Experimental settings}
For each dataset, we fine-tuned our proposed models for a maximum of 150 epochs using Adam optimizer with a learning rate of 5e-5, a weight decay of 0.1, and a batch size of 16 or 32. We selected the best model based on the performance in the validation set. For SPARQL query language, we set the maximum input length as either 128 (QALD, LC-QuAD, VQuAnDa) or 256 (Mon, Mon50, Mon80). We fixed the maximum length of natural language as 64. In the decoding step, we used beam search of beam width 10 for all the experiments. All experiments were completed using Python 3.7.10, Pytorch 1.8.1, and Transformers 4.5.1.

\subsection{Results}
\subsubsection{SPARQL Query Construction}
\def\arraystretch{1.2}% 
\setlength{\tabcolsep}{0.9em} %

\begin{table}[htbp]
\centering
\caption{Experiment results on SPARQL Query Construction}
\begin{adjustbox}{max width=\textwidth}
  \begin{tabular}{p{3cm}|c|c|c|c|c}
\hline
\centering \textbf{Model} & \textbf{QALD} & \textbf{Mon} & \textbf{Mon50} & \textbf{Mon80} & \textbf{LC-QuAD}  \\

\hline

RNN-Luong & 31.77$|$5.33 & 91.67$|$76 & 94.75$|$85.38 & 96.12$|$89.93 & 51.06$|$1.20 \\
Transformer & 33.77$|$6.00 & 95.31$|$91 & 93.92$|$84.70 & 94.87$|$85.80 & 57.43$|$7.80 \\
ConvS2S & 31.81$|$5.33 & 97.12$|$95 & \textbf{96.62}$|$90.91 & 96.47$|$90.87 & 59.54$|$8.20 \\
\hline
\hline
BERT2RND & 34.36$|$\textbf{6.67} & 97.03$|$96 & 95.28$|$90.69 & 96.44$|$91.35 & 66.52$|$14.80 \\
BERT2BERT & 35.86$|$\textbf{6.67} & 97.03$|$96 & 96.20$|$90.99 & 96.14$|$91.35 & 68.80$|$18.00 \\
\hline
\hline

\textbf{MLM} &&&&& \\
BERT2SPBERT (S) & 35.19$|$\textbf{6.67} & 97.28$|$\textbf{97} & 96.06$|$90.80 & 96.29$|$92.22 & 64.18$|$12.40 \\
BERT2SPBERT (B) & 35.95$|$\textbf{6.67} & 96.78$|$95 & 96.45$|$\textbf{91.18} & \textbf{96.87$|$92.70} & 68.08$|$\textbf{20.20} \\

\textbf{MLM + WSO} &&&&&\\
BERT2SPERT (B) &  \textbf{37.58$|$6.67} & \textbf{97.33}$|$96 & 96.20$|$90.84 & 96.36$|$91.75 & \textbf{69.03}$|$18.80 \\

\hline
\end{tabular}
\end{adjustbox}
\label{sparql_benchmark}
\textit{Notes:}  We train SPBERT (third group) from scratch (S) or initialized with the parameters of BERT (B), and we also use different learning objectives (only MLM, the combination of MLM and WSO). The left scores are BLEU and the right scores are EM. The best scores are in bold.
\end{table}
Table \ref{sparql_benchmark} shows the BLEU and EM results on the five datasets. Our approaches perform pretty well compared to the previous works. On the LC-QuAD, BERT2SPBERT\textsubscript{MLM+WSO}(B) achieves 69.03\%, creating new state-of-the-art results on this dataset. Moreover, one can see that BERT2SPERT\textsubscript{MLM+WSO}(B) outperforms the other models on the QALD dataset and improves by 3.81\% over the baselines, indicating our model can perform well even with limited data. SPBERT also obtains competitive results on three simple Mon datasets, including Mon, Mon50, and Mon80. BERT-initialized as an encoder can increase significant performance. However, the difference between BERT2RND and BERT2BERT is only apparent on QALD and LC-QuAD. They are almost equal on the three remaining datasets.
According to Table \ref{sparql_benchmark}, SPBERT outperforms the baselines in EM metric and is slightly better than BERT2BERT on some datasets. Specifically, BERT2SPERT\textsubscript{MLM}(B) achieves state-of-the-art results by more than 1\% on two out of five datasets. These results show that pre-training on SPARQL queries improves the construction of complete and valid queries.

\subsubsection{Answer Verbalization Generation}

In Table \ref{answer_table}, the results show that our proposed models significantly outperform BERT-initialized as encoder and the baseline methods for both \emph{With Entities} and \emph{Covered Entities}. Fine-tuning SPBERT improves results on the test set by 17.69\% (\emph{With Entities}) and 8.18\% (\emph{Covered entities}). The results also indicate that combining MLM and WSO can perform better than MLM when the input data are the SPARQL queries. At the same time, BERT2BERT is much better than BERT2RND in \emph{With Entities} setting, but BERT2BERT fails to predict in the sentences (\emph{Covered Entities}) that have no context and are incomplete.
\def\arraystretch{1.2}% 
\setlength{\tabcolsep}{0.9em} %

\begin{table}[htbp]
\centering
\caption{BLEU score experiment results on Answer Verbalization Generation}
\begin{tabular}{p{4cm}|P{1.5cm}|P{1cm}|P{1.5cm}|P{1cm}}
\hline
\multirow{2}{0pt}{\centering\textbf{Model}} & \multicolumn{2}{c|}{\textbf{With Entities}} & \multicolumn{2}{c}{\textbf{Covered Entities}}  \\
\cline{2-5}
 & Validation & Test & Validation & Test  \\
\hline

RNN-Luong & 22.29 & 21.33 & 34.34 & 30.78 \\
Transformer & 24.16 & 22.98 & 31.65 & 29.14 \\
ConvS2S & 26.02 & 25.95 & 32.61 & 32.39 \\
\hline
\hline
BERT2RND & 33.30 & 33.79 & 42.19 & 38.85 \\
BERT2BERT & 41.48 & 41.67 & 41.75 & 38.42 \\
\hline
\hline

\textbf{MLM} &&&& \\
- SPBERT2RND (S) & 43.25 & \textbf{43.64} & 41.49 & 38.57  \\
- SPBERT2RND (B) & 42.21 & 41.59 & \textbf{42.44} & 39.63 \\
- SPBERT2BERT (S) & 42.58 & 41.44 & 41.39 & 38.56 \\
- SPBERT2BERT (B) & 41.75 & 40.77 & 42.01 & 39.59 \\

\textbf{MLM + WSO} &&&&\\
- SPBERT2RND (B) & \textbf{43.52} & 42.39 & 41.74 & 38.48 \\
- SPBERT2BERT (B) & 43.23 & 41.70 & 41.97 & \textbf{40.57} \\

\hline
\end{tabular}
\label{answer_table}

\end{table}

\subsection{Discussion}
The experimental results show that leveraging pre-trained models is highly effective. Our proposed models are superior to baseline models on both tasks. The main reason for this is that we employ robust architectures that can understand the language from a bidirectional perspective. These models train on massive corpora, learning a generalized representation of the language that can be fine-tuned on smaller evaluation datasets.

The performance between the decoder initialized from the BERT checkpoint (BERT2BERT) and SPBERT checkpoint (BERT2SPBERT) is not significantly different in most cases. One possible reason is that we still need to randomly initialize the weights ($\sim$28M) for the attention mechanism between the encoder and the decoder, regardless of the decoder weights. This attention is used to align relevant information between the input and the output. Furthermore, we believe that this is why our proposed models underperform on SPARQL Query Construction compared to other tasks.

In Table \ref{sparql_benchmark}, our proposed models performed very well with up to 97.33\% BLEU score and 92.70\% EM score on some datasets such as Mon, Mon50, Mon80. On the other hand, QALD and LC-QuAD only obtained 37.58\% and 69.03\% BLEU scores while achieving 6.67\% and 20.20\% EM scores. One possible reason for this significant disparity comes from creating the evaluation datasets and the number of samples in these datasets. As previously mentioned in Section \ref{dataset}, QALD and LC-QuAD contain many complex question-query pairs which are manually constructed by a human. Meanwhile, the Mon dataset is generated automatically and lacks a variety of entities even though this dataset holds many samples.
In Table \ref{answer_table}, our models drop the performance with \emph{Covered Entities} setting when compared to \emph{With Entities} setting. That is because SPBERT is trained on executable SPARQL queries that fully contain entities and their relationships.

Results in the SPARQL Query Construction task are much higher than results in the Answer Verbalization Generation. Although VQuAnDa is the extension of LC-QuAD by expanding verbalized answers, BERT2SPBERT\textsubscript{MLM+WSO}(B) achieved 69.03\% BLEU score in LC-QuAD and SPBERT2BERT\textsubscript{MLM+WSO}(B) only obtained 40.57\% BLEU score in VQuAnDa. The possible reason is SPARQL is a structured query language, which almost starts with \texttt{SELECT} keyword and ends up with \texttt{brack\_close} (curly brackets). Meanwhile, natural answers are highly diverse, with many different sentences in the same meaning.

\section{Conclusion}
In this paper, we have employed pre-trained language models within a sequence-to-sequence architecture. We have also introduced SPBERT, which is the first structured pre-trained model for SPARQL query language. We conducted extensive experiments to investigate the effectiveness of SPBERT on two essential tasks of an end-to-end KBQA system. SPBERT obtains competitive results on several datasets of SPARQL Query Construction. The experimental results show that leveraging weights learned on large-scale corpora can outperform baseline methods in SPARQL Query Construction and Answer Verbalization Generation. SPBERT has also demonstrated that pre-training on SPARQL queries achieves a significant improvement in Answer Verbalization Generation task performance.

The number of our SPARQL queries is currently negligible compared to the number of triples (relationships) in the knowledge base. To improve the performance, we plan to expand the pre-training corpus. In addition, we can use an end-to-end architecture to improve efficiency. We plan to train a single BERT-SPBERT-BERT model on the two essential tasks rather than using BERT2SPBERT and SPBERT2BERT separately.

\bibliographystyle{splncs04}
\bibliography{iconip_2021}
%

% \begin{subappendices}
% \renewcommand{\thesection}{\Alph{section}}%
% % or try \arabic{section}

% \section{Pre-training Procedure}

% In the pre-training step, we denote the number of Transformer encoder layers as L, the size of hidden vectors as H, and the number of self-attention heads as A. We followed the setting of BERT\textsubscript{BASE} (L=12, H=768, A=12, total parameters=110M) and continued to train 200K steps from cased BERT\textsubscript{BASE} checkpoint. The maximum sequence length was fixed to 512, and the batch size was set to 128. It takes nearly 4 days to pre-train SPBERT. In addition, we used Adam with a learning rate of 2e-5 and epsilon of 1e-8 and employed cased BERT\textsubscript{BASE} vocabulary with 30K tokens. 

% \section{Fine-tuning Procedure}

% For each dataset, We fine-tuned our proposed models for a maximum of 150 epochs using a learning rate of 5e-5, a weight decay of 0.1, and a batch size of 16 and 32. We used Adam optimizer to update model parameters and selected the best model based on the performance in the validation set. For SPARQL query language, we set the maximum of length as 128 (QALD, LC-QuAD, VQuAnDa) and 256 (Mon, Mon50, Mon80). We fixed the maximum length of natural language as 64. For decoding, beam search of beam width 10 was used for all the experiments.

% All of the training used a single NVIDIA Tesla V100 GPU with 16GB memory to train SPBERT models and was completed using Python 3.7.10, Pytorch 1.8.1, and Transformers 4.5.1 installed.

% \end{subappendices}

\end{document}